\documentclass[5pt]{article}
\usepackage{hyperref}
\usepackage{times}
\usepackage[letterpaper]{geometry}
\usepackage{spconf,amsmath,epsfig}
\usepackage{enumitem}

\usepackage{fancyhdr}
\begin{document}\sloppy

\def\x{{\mathbf x}}
\def\L{{\cal L}}

\title{Perception Driven Texture Generation}
%
\name{Yanhai Gan$^\dagger$, Huifang Chi$^\dagger$, Ying Gao$^\dagger$, Jun Liu$^\ast$, Guoqiang Zhong$^\dagger$, Junyu Dong$^\dagger$\thanks{Thanks to XYZ agency for funding.}}
\address{$\dagger$Ocean University of China, Qingdao, China \\
$\ast$Qingdao Agricultural University, Qingdao, China\\
\{ganyanhai, chihuifang, gaoying\}@stu.ouc.edu.cn, liujunqd@163.com, \{gqzhong, dongjunyu\}@ouc.edu.cn}
%
%
%

\maketitle

\begin{abstract}
This paper investigates a novel task of generating texture images from perceptual descriptions. Previous work on texture generation focused on either synthesis from examples or generation from procedural models. Generating textures from perceptual attributes have not been well studied yet. Meanwhile, perceptual attributes, such as directionality, regularity and roughness are important factors for human observers to describe a texture. In this paper, we propose a joint deep network model that combines adversarial training and perceptual feature regression for texture generation, while only random noise and user-defined perceptual attributes are required as input. In this model, a preliminary trained convolutional neural network is essentially integrated with the adversarial framework, which can drive the generated textures to possess given perceptual attributes. An important aspect of the proposed model is that, if we change one of the input perceptual features, the corresponding appearance of the generated textures will also be changed. We design several experiments to validate the effectiveness of the proposed method. The results show that the proposed method can produce high quality texture images with desired perceptual properties.
\end{abstract}
\begin{keywords}
Texture Generation, Perceptual Features, Adversarial Training, Regression, Neural Networks
\end{keywords}
\section{Introduction}
\label{sec:intro}
Textures play important roles in multimedia applications, such as understanding and generation of multimedia content. Texture synthesis and generation have also been extensively investigated in the past years~\cite{Badalov2011An}. Before the revival of deep learning, researchers mainly used example-based approaches to synthesize textures. In these methods, new textures with similar appearances to existing samples can be produced. With the development of deep learning, more methods for texture generation have been proposed by learning from the training data. However, there is no visual perceptual information involved in this process, whereas humans commonly use perceptual attributes, such as texture roughness, coarseness and directionality, to describe textures. Moreover, the majority of deep learning based methods can only generate images of low quality. Thus, it is desired to develop a new way for generating high-quality textures based on human perceptual descriptions; for example, the new generation method should be able to produce textures with strong directionality or less regularity as required by the user.

Convolutional Neural Network (CNN), which was inspired by the mechanism of visual cortex, has shown great superiority in latest studies~\cite{jarrett2009best}~\cite{Krizhevsky2012ImageNet}. With the aid of deep convolutional networks, researchers have made breakthroughs in many classical computer vision tasks. For example, in the ImageNet Large Scale Visual Recognition Challenge, the performance of computer algorithms even surpassed human's~\cite{szegedy2015rethinking}. Consequently, researchers have been investigating different approaches based on CNN for image generation~\cite{hinton2006fast}~\cite{gregor2015draw}~\cite{dosovitskiy2015learning}.

Goodfellow et al. proposed a generative adversarial framework (GAN)~\cite{goodfellow2014generative} and produced excellent results in many image generation tasks. However, the generated samples were still in low resolution and far from being perfect. In order to generate more realistic images, Wang and Gupta factorized the image generation process and proposed a joint model consisting of Style and Structure Generative Adversarial Networks~\cite{wang2016generative}. Experimental results in ~\cite{wang2016generative} suggested that a great gain could be obtained through this factoring trick for generating realistic indoor scenes. All these work indicates that it is a promising practice to exploit joint convolutional neural networks and adversarial training schemes for generating high-quality images.

In addition to generating natural images, another question is what we can generate from semantics or high-level descriptions. Many efforts have been made regarding this topic. Karpathy et al. proposed a fragment embedding method in 2014~\cite{karpathy2014deep}, which was essentially a bidirectional retrieval scheme, as the desired image must exist in the image database. Yan et al. modeled images as composite of foreground and background and developed a layered generative model~\cite{yan2015attribute2image}. Their method shows promising results in the tasks of attribute-conditioned image reconstruction and completion. Nevertheless, the quality of generated images is still not good enough for texture perception study.

The contribution of this paper is a new joint model that combines perceptual feature regression and adversarial schemes for generating textures based on perceptual descriptions. Unlike existing Conditional Generative Adversarial Networks (CGAN)~\cite{mirza2014conditional}, in which the discriminative model need to estimate the joint distribution of condition vectors with samples and can not always provide enough information for the generator to adjust parameters, in our new model, perceptual feature regression can supervise the generator to produce textures in consistence with human visual system. Thus, the discriminative model is assisted by the perceptual regression model and therefore released from the inaccurate estimation of joint distributions. Furthermore, the perceptual model is able to supply more information to the generator and guide it to produce texture with enough details, which lead to high-quality output texture images.

\section{Related Work}

Textures have attracted widespread attention in the research field of visual perception and computer vision. Rao et al. identified the perceptual features people used to classify the textures and also established the correlation between semantic attributes and textures~\cite{bhushan1997texture}, which showed the importance of perceptual features for understanding texture images. Meanwhile, texture synthesis and texture generation have been active research areas for many years. Shin et al. proposed a pixel-based method for texture synthesis with non-parametric sampling~\cite{shin2006pixel}, and Wei proposed an efficient algorithm using tree-structured vector quantization for realistic texture synthesis, which required only a sample texture as input~\cite{wei2000fast}. These studies normally concern on example based texture synthesis, whereas our work focuses on generating textures according to user-defined perceptual attributes.

Deep learning models, particularly deep convolutional neural networks, have achieved great success in texture analysis due to their strong learning capability. Texture synthesis based on CNN is a new research topic~\cite{gatys2015texture}, which has produced promising results. These results suggest that this topic deserves more research devotion. In ~\cite{gatys2015texture}, Gatys combined the conceptual framework of spatial summary statistics on feature responses with the feature space of a convolutional neural network, and the goal is to generate textures from a given source image. Ulyanov also trained feed-forward generation networks to generate multiple samples of the same texture with arbitrary sizes~\cite{ulyanov2016texture}. In this manner, the representation of the given image can be learned by the convolutional networks, and the new samples can be generated from the networks. Goodfellow~\cite{goodfellow2014generative} proposed a generative adversarial framework that could estimate generative models via an adversarial process, in which a generative model $G$ and a discriminative model $D$ were simultaneously trained. The generative model is responsible for capturing the data distribution, and the discriminative model is used to estimate the probability that a sample comes from the training data rather than $G$. The training procedure for $G$ is to maximize the probability of $D$ making a mistake. It has been proven that GAN can be used to generate realistic images from uniformly distributed random noise~\cite{goodfellow2014generative}. Furthermore, GAN was extended as CGAN for conditional image generation by Mirza and Osindero~\cite{mirza2014conditional}, where both models $G$ and $D$ received an additional vector of information as condition. This vector might contain information about the class of the training example. CGAN has been successfully applied in digit and face image generation~\cite{gauthier2014conditional}, whereas we are interested in generating textures with given perceptual attributes.

Inspired by previous works, this paper proposes a joint model, which combines the perceptual feature regression and adversarial training scheme for perception driven texture generation. Since the perceptual regression model can provide additional information for the generator in the adversarial scheme, the proposed model is able to generate high-quality textures.

\section{Perception Driven Texture Generation}

In this section, we first introduce the overall architecture of the proposed joint model for perception driven texture generation. Then we provide details on the network design and initialization.

\subsection{Overall Architecture of the Joint Model}

\begin{figure*}
  \centering
  \includegraphics[width=17cm]{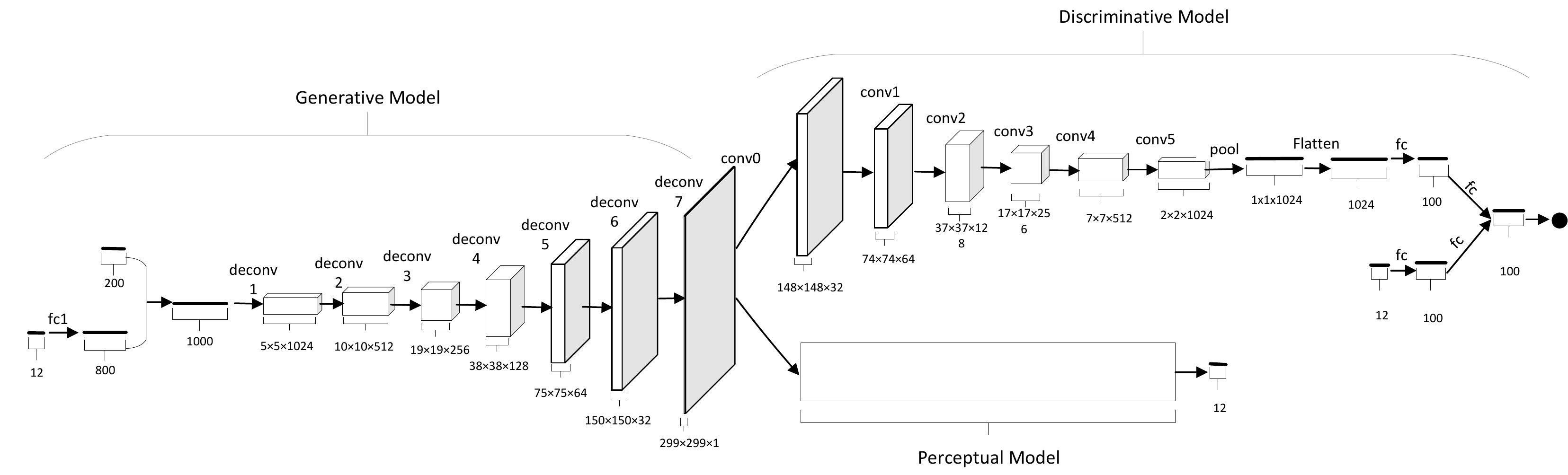}\\
  \caption{\textbf{Architecture of the joint models for perception driven texture generation.}}\label{fig:4}
\end{figure*}

Human observers essentially use perceptual features for texture description, e.g. regularity and repetitiveness~\cite{Rao1996Towards}. According to ~\cite{liu2015visual}, there are 12 prominent perceptual features for human to perceive a texture. In practice, human can not only perceive these features from a texture but also imagine a texture from these perceptual descriptions. For example, textures with weak or strong directionality can be easily depicted in human mind; in contrast, no computer algorithm is able to generate texture from these descriptions. Therefore we designed a joint deep model in order to achieve such a goal. As shown in Fig.~\ref{fig:4}, the overall architecture includes three parts: a perceptual feature regression model, a conditional generative model, and a discriminative model. The generative model is responsible for conditional texture generation, whereas the discriminative model is used to distinguish whether the generated texture is from the training sample distribution, and the perceptual model can drive the generative model to produce textures possessing certain attributes.

Inspired by the success of the Inception-v3 model~\cite{szegedy2015rethinking}, which reached 3.46\% top-5 error rate and even surpassed human performance in the 2015 ImageNet Large Scale Visual Recognition Challenge(ILSVRC), we use Inception-v3 for our perceptual feature regression. First we change the activation function of the final output layer and auxiliary units to $tanh$, as our perceptual features are scaled in the range between -0.9 and 0.9. The reason for scaling the range is to avoid the saturation of the output neurons. Furthermore, $tanh$ is much easier to be trained than $sigmoid$~\cite{glorot2010understanding}. Second, we change the cross entropy loss of softmax to the quadratic loss. Then we train the modified Inception-v3 model using our texture database for perceptual feature prediction. In the following sections, we call the modified Inception-v3 as the perceptual model.

In the CGAN framework, the discriminator needs to figure out the union distribution of the condition and samples. The distinguishing task is relatively difficult, and the discriminator cannot supply enough information for the generator to justify its parameters. In our model, we use the perceptual model to impose perceptual constraints on the generator; this can provide additional information for the generator to produce certain perceived textures.
We use $G$, $D$, and $H$ to represent the generative, discriminative and perceptual model, respectively. Then the loss of $D$ can be defined as:
\begin{equation}\label{eq:9}
  D\_loss=-\frac{1}{n}\sum_{i=1}^{n}(q_i\ln(D(x_i,y_i))+(1-q_i)\ln(1-D(x_i,y_i))),
\end{equation}
where $x_i$ represents a training example, $y_i$ is the corresponding perceptual feature vector, $q_i$ is one or zero, indicating whether $(x_i,y_i)$ is a real pair, and $n$ is the number of training examples. The quadratic loss for $H$ is defined as:
\begin{equation}\label{eq:10}
  H\_loss=\frac{1}{2n}\sum_{i=1}^{n}(H(x_i)-y_i)^{2}.
\end{equation}
The loss of $G$ contains two parts: one from $D$, and the other from $H$; the definition is:
\begin{eqnarray}\label{eq:11}
  G\_loss&=&G\_loss\_d+\alpha* G\_loss\_h\\
  G\_loss\_d&=&-\frac{1}{n}\sum_{i=1}^{n}\ln(D(G(y_i,z_i),y_i)) \\
  G\_loss\_h&=&\frac{1}{2n}\sum_{i=1}^{n}(H(G(y_i,z_i))-y_i)^{2},
\end{eqnarray}
where $\alpha$ is a tradeoff parameter, $z_i$ is a random noise vector, $H$ is preliminarily trained, and $G$ and $D$ are trained in an adversarial scheme. In this manner, the discriminator makes the generator produce realistic textures, and the perceptual model makes the generated textures possess certain perceptual attributes.
\subsection{Network Design Details}

In this subsection, we first introduce the initialization scheme for our deep networks, and then present strategies for the design of certain part of the network. Inspired by~\cite{he2015delving}, we initialize weights of one layer of the proposed network by formulation $Var[w]=2/n$. In most cases, we only consider the back propagation situation, so $n$ represents the number of units that can be reached by one input neuron, and $w$ represents the weight in convolutional or fully connected layer. ReLU is used as the activation function in the network, since it can reduce the gradient vanishing effect and make the model learn fast. However, we would like the output of the generator to be limited in a certain range, because an image always has limited pixel values. The discriminator should yield a probability result, which indicates whether an image comes from the real training samples. Accordingly, we use $tanh$ as the activation function in the output layer of the generative model, and $sigmoid$ in the discriminative model. Thus, we adopt different initialization strategies for the output layer. In order to keep the gradient variance, when the activation function is $tanh$, we initialize the weights using the truncated normal distribution with the standard deviation $\sqrt{1/n}$. In contrast, we use $4\sqrt{1/n}$ as the deviation when the activation function is $sigmoid$. Here, we assume that the weights are initialized independently, and the bias is initialized with zero. In particular, if the number of units decreases too much in the output layer, we slightly reduce the deviation of weights to avoid the output becoming too saturated in the forward case. We will introduce more details about the network design in section 4.

It should be noted that, in the fully connected layer, the initialization strategy can be easily analyzed. However, it becomes complicated in the convolutional layers. We may take the 1-D convolutional operation as an example, and it can be easily extended to the high dimensional case. We use $n$ to represent the number of units, which can propagate its gradient to certain input unit. When the number of input units becomes very large, we can calculate an average value for $n$. We define a universal formulation:
\begin{eqnarray}\label{eq:5}
  &\lfloor\frac{k-1}{d}\rfloor+1& \\ \nonumber
  &\lfloor\frac{k}{d}\rfloor& \\ \nonumber
  &\lfloor\frac{k+1}{d}\rfloor& \\ \nonumber
  &\vdots& \\ \nonumber
  &\lfloor\frac{k+d-2}{d}\rfloor,&
\end{eqnarray}
where $\lfloor x\rfloor$ represents the maximal integer no larger than $x$, $k$ represents the kernel size, and $d$ represents the step size. Eq~\eqref{eq:5} illustrates a period of the convolutional operation. Each line in Eq~\eqref{eq:5} calculates the number of units that can be reached by certain input unit. The period begins with the $kth$ input unit. The length of the cycle is $d$. From Eq~\eqref{eq:5}, we can get the average value of $n$ for general situation:
\begin{equation}\label{eq:6}
  \frac{\sum_{i=0}^{d-2}\lfloor\frac{k+i}{d}\rfloor+\lfloor\frac{k-1}{d}\rfloor+1}{d}.
\end{equation}
We use the average value of $n$ to calculate the deviation of $w$ for initialization. To extend this to the two dimensional situation, we simply expand $k$ and $d$ to two dimensions. This scheme is used to initialize our networks through all experiments.

In order to emphasize the importance of perceptual features for texture generation, we stretch the perceptual feature vector to 800 dimensions via a fully connected layer. The random noise vector is drawn uniformly from a 200 dimensional space ranging from -1 to 1. The reason for using these specific dimensions is explained as follows.
A random noise vector with 200 dimensions can be significantly varied to generate diverse textures given certain perceptual features. In theory, if we change each dimension of the random noise vector with step of 0.1, we can obtain $20^{200}$ different vectors. This is a large enough space for variant texture appearance. In addition, textures with the same perceptual feature vector have similar appearances. In the above analysis, we demonstrate that the covariance shift can be avoided by certain initialization strategy in the forward and backward view. In the fully connected layers for stretching perceptual features, we simply consider the forward propagation. Thus, we make $n$ represent the number of units in the input layer. Consequently, the stretched perceptual features own similar variance as the original. Let $z$ represent the random variable. Then its variance is $Var[z]=1/3$. Recall that the perceptual features are scaled to the range between -0.9 and 0.9. Let $f$ represent one perceptual feature, and we use the following equation for scaling:
 \begin{equation}\label{eq:8}
   \hat{f}=min(max((f-E(f))/\sigma(f),-3),3)\times0.3.
 \end{equation}
 Through this transformation, the resulted $\hat{f}$ owns variance of $0.09$. Since the stretching layer is initialized by using the forward principle, the variance of the stretched features is also approximately $0.09$. The result is that the variance of the random noise is three times larger than that of the stretched perceptual features. Hence if we want the perceptual features to play the same role as random noise in the generating task, we should make the number of the output units in the stretching layer three times larger than that of the random noise. In this work we therefore set the number to 800, and we can let the perceptual features dominate the generating procedure.

\section{Experiments}
\subsection{The Data Set}

In our experiments, we use the Perceptual Texture Database (PTD), in which there are 450 textures with corresponding 12-D perceptual features~\cite{liu2015visual}. The textures in PTD have a resolution of $512\times512$, and the 12-D perceptual features include contrast, repetitiveness, granularity, randomness, roughness, density, directionality, structural complexity, coarseness, regularity, orientation and uniformity. However, since 450 textures are still too few to train a deep neural network, we expand the examples in the following way. First, we crop each texture into 81 textures of size $448\times448$; the step used for cropping is 8. Second, we resize the resulted textures to $299\times299$. Regarding perceptual features, we let the resulted textures have same values as their original ones. We eventually obtain 36450 examples of size $299\times299$, and we use 36000 among them to train our models. The remaining textures are left as the validation set. It should be noted that it is reasonable to make the resulted 81 textures have the same perceptual features as their originals. First, the textures in PTD are isotropic; a $448\times448$ region can cover most area of the original texture and can therefor keep original perceptual characteristics. Second, resizing the $448\times448$ texture to $299\times299$ does not cause obviously blurring effect.

\subsection{Perceptual Feature Regression}

\begin{figure}
  \centering
  \includegraphics[width=7cm]{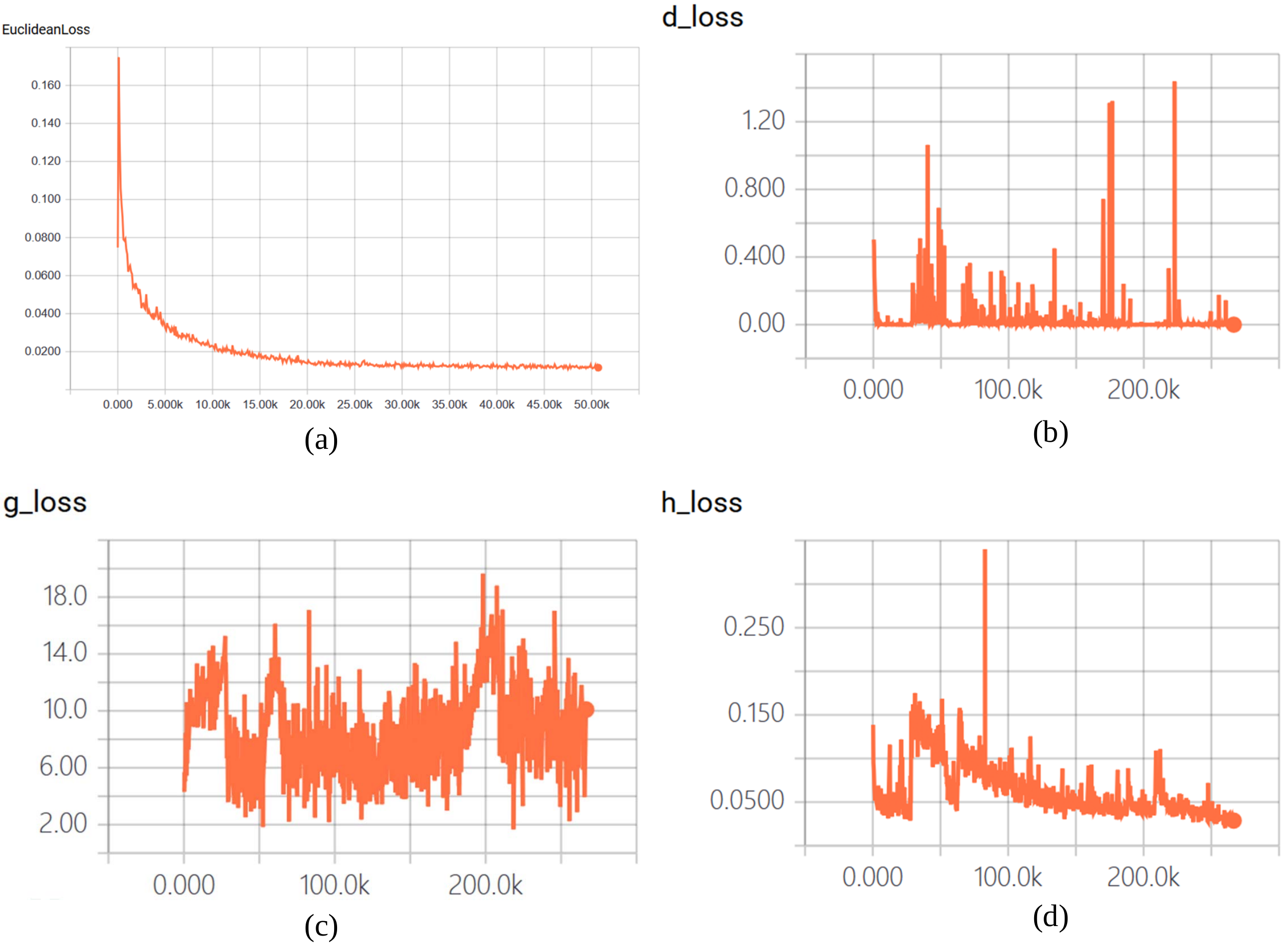}\\
  \caption{\textbf{Training loss curves of the jointed models.} (a) is the regression curve. (b) is the discriminative model training loss curve. (c) is the generative model training loss curve. (d) is the perceptual model training loss curve.}\label{fig:step4_curves}
\end{figure}

Since our perceptual model was modified from Inception-v3, we did not need to train it from scratch. The preliminary trained Inception-v3 on ImageNet can be found in ~\cite{inception-v3}. Since our perceptual model only differed from Inception-v3 in the output layer and loss definition, we initialized the output layer with truncated Gaussian noise, and the other layers were reloaded from preliminary trained Inception-v3 model. Then we fine-tuned the perceptual model with initial learning rate 0.001. The RMSProp method was used for gradient descent~\cite{overview}. We ran the optimization algorithm for 50000 iterations. The process is illustrated in Fig.~\ref{fig:step4_curves}(a). Finally, the Euclidean loss converged to 0.01161, and the final evaluation error was 0.0039. Since the perceptual features have 12 attributes, the standard error deviation for each attribute in average can be calculated:
\begin{equation}\label{eq:12}
  \sigma(e)=\sqrt{0.01161\times2/12}=0.044.
\end{equation}
This means that we can accurately predict the perceptual features for one texture with very small deviation. Based on this observation, we can make a basic assumption here: if the generated textures have certain perceptual attributes, it should be correctly perceived by the perceptual model. We use the preliminary trained perceptual model as an accessory of the whole generative framework.

\subsection{Generating Textures from Perceptual Features}

To generate realistic textures, we must design a reasonable network structure. The kernel size is a vital factor for generating high-quality images. In the experiments, we found that if we set the kernel size too small, i.e. 3, the generated textures owned more details but looked too crude. If the kernel size was too large, i.e. 7, the generated textures looked more smooth, but with less details. Eventually, we used $5\times5$ kernels for convolution or inverse convolution in our discriminative and generative models. We also tried to fuse kernels of different size for generating textures with more details and global information. However, it did not produce good results.

\begin{figure}
  \centering
  \includegraphics[width=8cm]{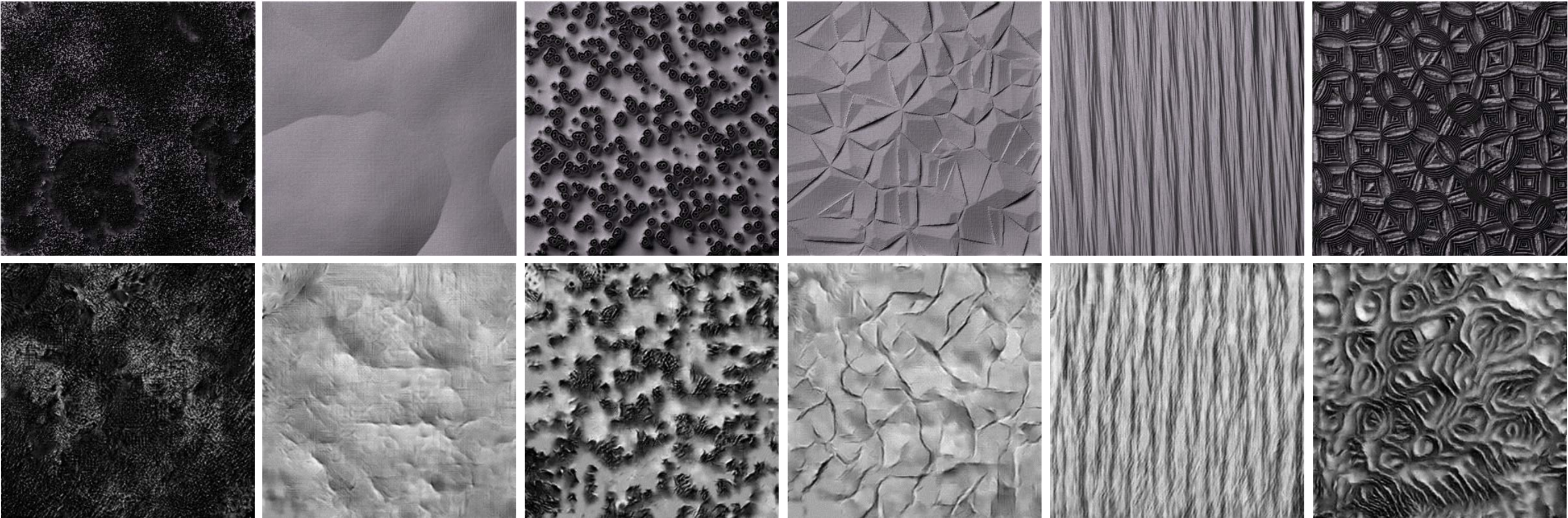}\\
  \caption{\textbf{Textures generated from existing perceptual feature vectors.} Textures in the same column have the same perceptual features. Textures in the first row are from our database, and those in the second row are generated from corresponding perceptual feature vectors.}\label{fig:step4_generated_textures}
\end{figure}

\begin{figure}
  \centering
  \includegraphics[width=8cm]{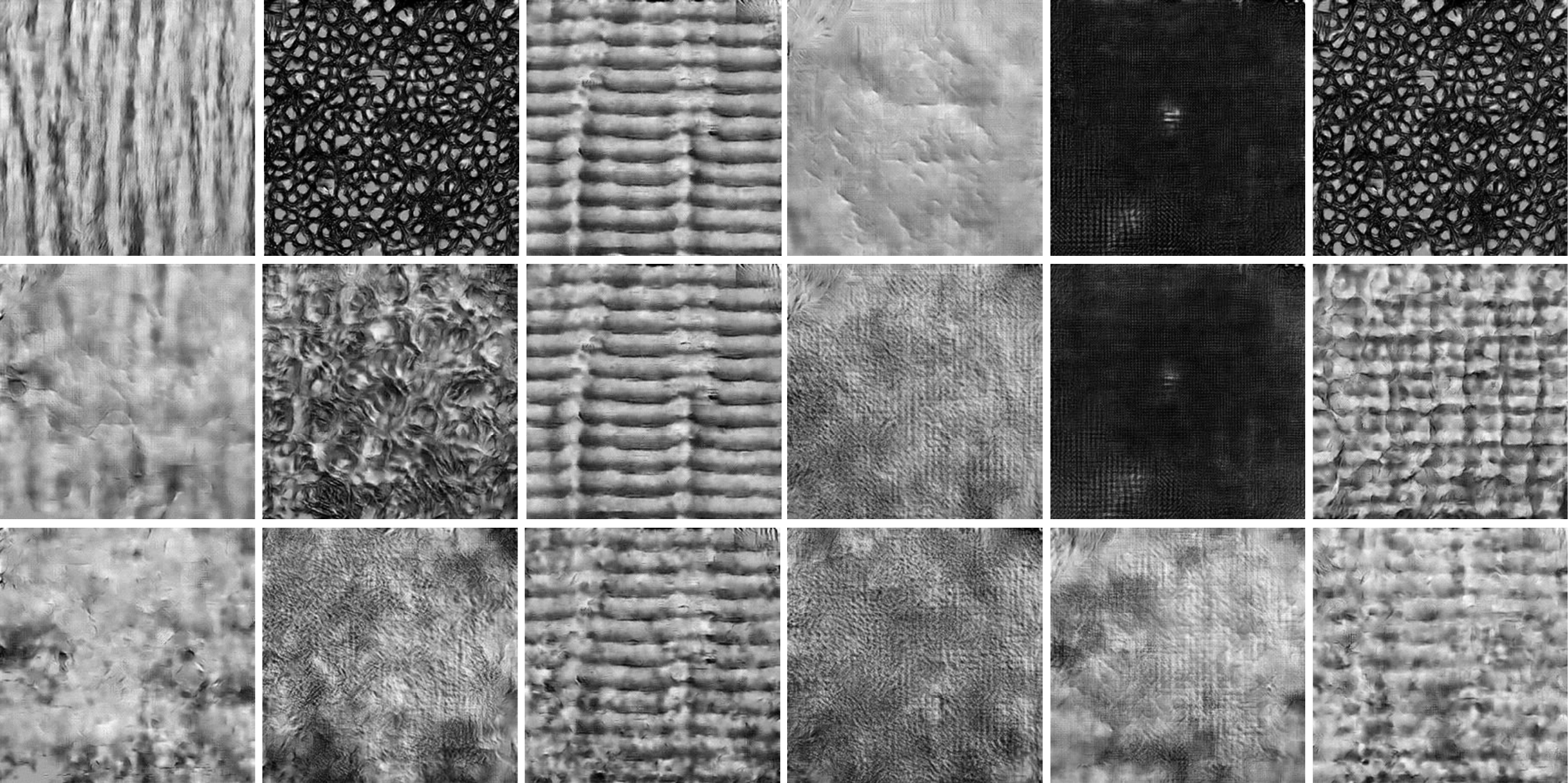}\\
  \caption{\textbf{Textures generated from manually created perceptual features.} In each column, we vary one perceptual feature from large to small in \{0.9, 0, -0.9\}, while the others are kept the same as the existing perceptual features. The following perceptual features are manually changed: directionality, contrast, granular, roughness, feature density, structural complexity. It can be seen that corresponding features gradually become less obvious.}\label{fig:step4_handcrafted_textures}
\end{figure}

Since one part of the input to the generative model was drawn from random noise (the other part is the perceptual feature vector), there were infinitely many training examples in practice. Thus we used the ADAM~\cite{kingma2014adam} method for optimization. We optimized the generative model twice after each optimization for the discriminative model. We made each batch contain 60 training examples. The tradeoff parameter $\alpha$ was set as 10. In the end, we ran 266000 optimization iterations. The training process is illustrated in Fig.~\ref{fig:step4_curves}(b)(c)(d). Two experiments were designed after the models were trained. First, we fed real perceptual features in our database with different random noise to the generative model. The generated textures are shown in Fig.~\ref{fig:step4_generated_textures}. Second, we manually edited some perceptual features and used them to generate textures. It should be emphasized that the manually edited or handcrafted perceptual features were based on existing perceptual features, i.e. only certain perceptual feature was set to three different values: 0.9, 0, -0.9, whereas the others were kept the same as the existing ones. In Fig.~\ref{fig:step4_handcrafted_textures}, we only provide six results due to the limited space, but more results are provided in the supplementary materials. As an example, we can see from the first column of  Fig.~\ref{fig:step4_handcrafted_textures}, when we decrease the perceptual feature value of directionality from 0.9 to -0.9, the textures gradually lose the overall direction. These results indicate that the proposed method is able to generate desired textures by varying certain perceptual attributes.

\section{Conclusion}
We propose a novel deep network model for perception driven texture generation. In the proposed model, a perceptual regression component is integrated with the generative framework, which drives the produced textures possessing certain perceptual attributes. This perceptual regression model partially releases the discriminative model's workload, and can supply more information for the generator to produce better perceived texture. Experimental results show that the jointed models are able to generate realistic texture from given perceptual attributes. We attribute this success to the fact that if the generated texture is realistic enough, it should have the potentiality to be correctly perceived by the preliminary trained deep network.

It should be noted that the perceptual features are not independent from each other. If we change one perceptual attribute arbitrarily, the remaining relevant features might also need to be changed to fit the real distribution. In the future work, we will design an auxiliary model for generating correct perceptual feature vectors; in this way we may simply provide an existing perceptual feature vector and the desired value for certain attribute, and the tool can generate a suitable input perceptual feature vector.

\bibliographystyle{IEEEbib}

\end{document}